\documentclass[conference]{IEEEtran}
\IEEEoverridecommandlockouts
\usepackage{cite}
\usepackage{amsmath,amssymb,amsfonts}
\usepackage{graphicx}
\usepackage{textcomp}
\usepackage{xcolor}
\def\BibTeX{{\rm B\kern-.05em{\sc i\kern-.025em b}\kern-.08em
    T\kern-.1667em\lower.7ex\hbox{E}\kern-.125emX}}

\usepackage{tikz}
\usepackage{calc}

\usepackage[noabbrev]{cleveref}

\usepackage{subfigure}
\usepackage{booktabs} 
\usepackage{caption}
\usepackage{stfloats}
\usepackage{wrapfig}

\usepackage{pifont}
\newcommand\mynuma[1]{\ifcase#1 \or \ding{172}\or \ding{173}\or
  \ding{174}\or \ding{175}\or \ding{176}\or \ding{177}%
  \or \ding{178}\or \ding{179}\or \ding{180}\or \ding{181}\else *\fi\relax}

\newcommand\mynumb[1]{\ifcase#1 \or \ding{182}\or \ding{183}\or
  \ding{184}\or \ding{185}\or \ding{186}\or \ding{187}%
  \or \ding{188}\or \ding{189}\or \ding{190}\or \ding{191}\else *\fi\relax}

\usepackage{nccmath}  
\usepackage{bbm}
\usepackage{amsmath}
\usepackage{amssymb}
\usepackage{mathtools}
\usepackage{amsthm}
\usepackage{multirow}
\usepackage{booktabs}
\usepackage{xcolor}
\usepackage{booktabs}
\usepackage{enumitem}

\usepackage[titlenumbered,ruled]{algorithm2e}
\usepackage{algorithmic}
\usepackage{array}
\usepackage{wrapfig}
\usepackage{fdsymbol}

\usepackage{url}
\usepackage{tikz}
\usepackage{comment}
\usepackage{amsmath,amssymb} 
\usepackage{color}
\usepackage{microtype}
\usepackage{graphicx}
\usepackage{subfigure}
\usepackage{booktabs} 
\usepackage{ulem}
\usepackage{mathtools}
\usepackage{amsthm}
\usepackage{multicol}
\usepackage{multirow}
\usepackage{makecell}
\usepackage{stfloats}
\usepackage{wrapfig}
\usepackage{enumitem}
\usepackage{nccmath}

\definecolor{Note_color}{rgb}{0.0, 0.0, 1.0}

\begin{document}

\title{
NetBooster: Empowering Tiny Deep Learning By
Standing on the Shoulders of Deep Giants \vspace{-0.2em}
\thanks{
This work was supported by the NSF SCH program (Award
number: 1838873) and NSF NIH program (Award number: R01HL144683).}
\vspace{-0.2em}
}

\author{
\IEEEauthorblockN{Zhongzhi Yu$^{1}$, Yonggan Fu$^{1}$, Jiayi Yuan$^{2}$, Haoran You$^1$, Yingyan (Celine) Lin$^1$}
\IEEEauthorblockA{$^1$\textit{Georgia Institute of Technology}, $^2$\textit{Rice University}}
\IEEEauthorblockA{\textit{\{zyu401, yfu314, hyou37, celine.lin\}@gatech.edu, jy101@rice.edu}}
\vspace{-2.0em}
}

\maketitle

\begin{abstract}
Tiny deep learning has attracted increasing attention driven by the substantial demand for deploying deep learning on numerous intelligent Internet-of-Things devices. However, it is still challenging to unleash tiny deep learning's full potential on both large-scale datasets and downstream tasks due to the under-fitting issues caused by the limited model capacity of tiny neural networks (TNNs). To this end, we propose a framework called NetBooster to empower tiny deep learning by augmenting the architectures of TNNs via an expansion-then-contraction strategy. Extensive experiments show that NetBooster consistently outperforms state-of-the-art tiny deep learning solutions.
    
\end{abstract}

\section{Introduction}
\label{sec:intro}
Tiny deep learning, which aims to develop tiny neural networks (TNNs) featuring much-reduced network sizes along with lower memory and computational costs, has emerged as a promising direction to enable deep learning's wider real-world applications in resource-constrained Internet-of-Things (IoT) devices~\cite{lin2020mcunet} and has attracted an increasingly growing interest from both industry and academia~\cite{lin2020mcunet,cai2021network}. In particular, existing tiny deep learning works strive to improve the achievable accuracy-efficiency trade-off of TNNs by either designing novel efficient network architectures~\cite{lin2020mcunet,sandler2018mobilenetv2} or compressing a large deep neural network (DNN) to reduce their network redundancy~\cite{fu2021double}. However, the achieved accuracy-efficiency trade-off of existing TNN works is still far from satisfactory for many IoT emerging applications~\cite{cai2021network}. Specifically, we summarize that the constraints to the achievable accuracy-efficiency trade-offs of TNNs stem from two factors: \uline{\textbf{\textit{Constraint 1}}}: It is challenging for TNNs to learn complex but representative features~\cite{cai2021network} and achieve satisfactory accuracy on commonly used large-scale datasets (e.g., ImageNet), and \uline{\textbf{\textit{Constraint 2}}}: TNNs' limited accuracy on large-scale datasets further hinders TNN-based solutions from leveraging the widely adopted pretrain-then-finetune paradigm for real-world downstream tasks.

\begin{figure}[!ht]
    \centering
    \includegraphics[width=0.8\linewidth]{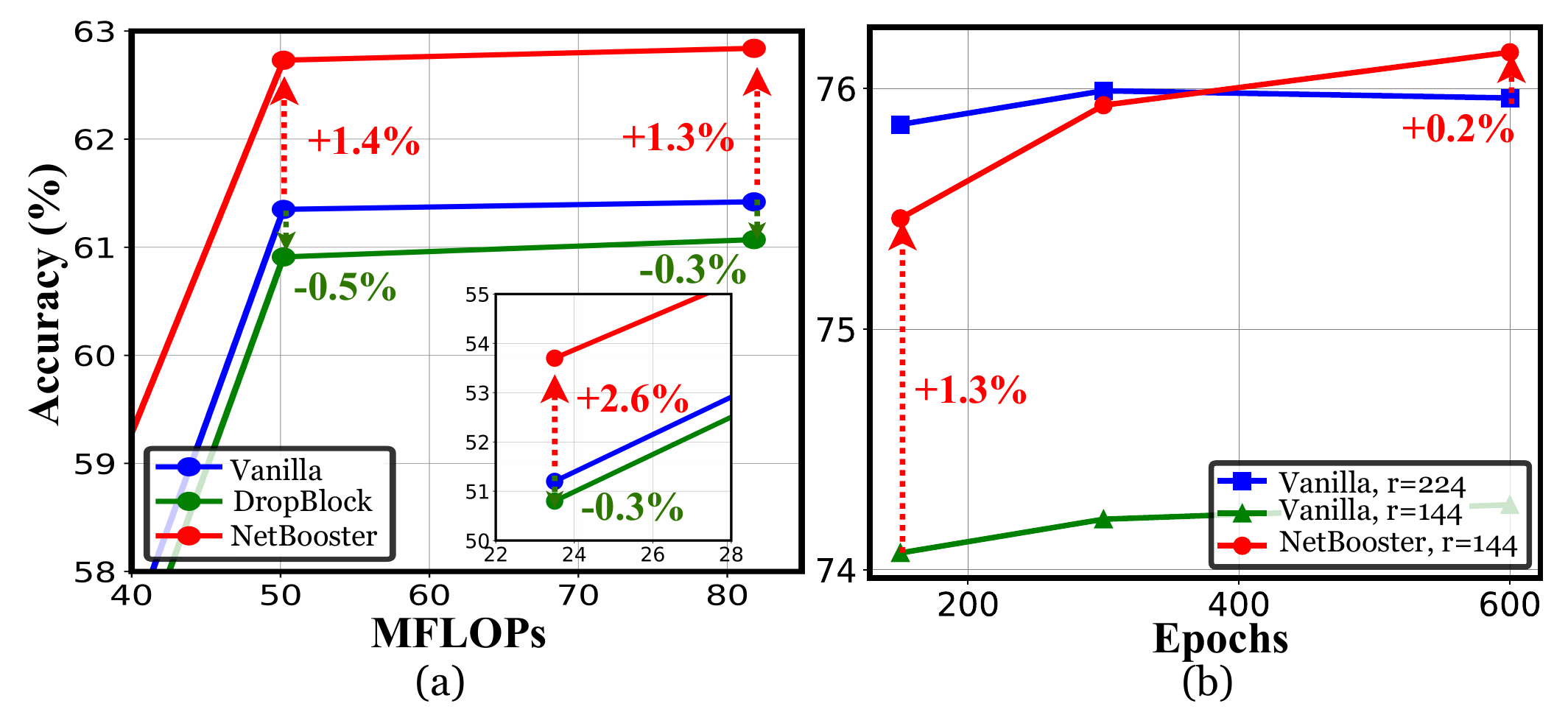}
    \vspace{-0.6em}
        \caption{(a) \uline{\textbf{\textit{Constraint 1}}}: TNN training suffers from under-fitting issues. When training MobileNetV2~\cite{sandler2018mobilenetv2} on ImageNet, regularization techniques (e.g., \textcolor{green}{DropBlock}~\cite{ghiasi2018dropblock}) even lead to inferior accuracy compared with \textcolor{blue}{vanilla training}. Our proposed \textcolor{red}{NetBooster} can boost TNNs' accuracy by increasing its capacity during training. (b) \uline{\textbf{\textit{Constraint 2}}}: Inadequately trained TNNs cannot learn complex features and thus suffer from limited downstream task accuracy. Finetuning ImageNet pretrained vanilla MobileNetV2-35 with a resolution of \textcolor{blue}{$224 \times 224$} and \textcolor{green}{$144 \times 144$}, respectively, on the CIFAR-100 dataset for even four times more epochs (i.e., 600 epochs) still cannot improve the achievable accuracy. Our proposed \textcolor{red}{NetBooster} can boost TNNs' accuracy by inheriting pretrained deep giants' learned complex features. 
        }
    \label{fig:fig1}
    \vspace{-1.5em}
\end{figure}

In parallel, it has recently been recognized that a dedicated training recipe can boost the accuracy of TNNs~\cite{cai2021network}, although this area remains still under-explored. Unlike DNN training, which requires techniques like data augmentation~\cite{cubuk2020randaugment} and/or regularization~\cite{ghiasi2018dropblock,srivastava2014dropout} to alleviate the \textit{over-fitting} issue, a recent study~\cite{cai2021network} has shown that the small network capacity of TNNs makes them more prone to the \textit{under-fitting} issue. 
Specifically, due to TNNs' limited ability to learn complex features, extensively augmented training data or a heavily regularized training process can hurt the achievable accuracy of TNNs on large-scale datasets \uline{(i.e., \textbf{\textit{Constraint 1}})}, as shown in Fig.~\ref{fig:fig1} (a). The lack of learned complex and representative features in the pretrained TNNs further limits the achievable accuracy of downstream tasks, which cannot be mitigated by additional training epochs \uline{(i.e., \textbf{\textit{Constraint 2}})}, as shown in Fig.~\ref{fig:fig1} (b).

To narrow the gap between the increasing demand for more powerful TNNs in real-world applications and the lack of effective TNN training schemes, we aim to develop a technique that can boost the achievable task accuracy of TNNs, while preserving their appealing efficiency, by empowering TNNs' learned features. In particular, this work makes the following contributions: 

\begin{itemize}
    \item[$\bullet$] To the best of our knowledge, we are the first to discover and promote a new paradigm of training TNNs to boost their achievable accuracy via constructing a competent deep giant using compound network augmentation (i.e., augmenting both width and depth dimensions of the given TNNs), which is simple, effective, and generally applicable.  

    \item[$\bullet$] By leveraging the above discovery, we propose a TNN training framework, dubbed NetBooster, that alleviates TNNs' under-fitting issue during training, boosting their achievable accuracy while preserving their original network complexity and thus inference efficiency. Specifically, NetBooster incorporates a two-step expansion-then-contraction training strategy: \textbf{Step-1: Network Expansion} constructs an expanded deep giant by converting some layers of the original TNN into multi-layer blocks, facilitating the learning of more complex features by leveraging the corresponding deep giant counterpart, which equips the original TNN with an initial state already possessing sufficient knowledge, and \textbf{Step-2: Progressive Linearization Tuning (PLT)} then reverts the deep giant back to the original TNN's structure by removing the non-linear layers from the expanded blocks and then contracting them.
    
    \item[$\bullet$] We make heuristic efforts to empirically investigate the optimal setting for effectively boosting the achievable accuracy of TNNs when implementing Network Expansion in NetBooster. Specifically, we address the following questions: $\mathcal{Q}1$. What kind of block to use for expansion, $\mathcal{Q}2$. Where to expand within a TNN, and $\mathcal{Q}3$. How to determine the expansion ratio.
    
    \item[$\bullet$] Extensive experiments and ablation studies on two tasks, four networks, and seven datasets demonstrate that NetBooster consistently achieves a non-trivial accuracy boost (e.g., 1.3\% $\sim$ 2.6\%) compared to state-of-the-art (SOTA) TNNs on the ImageNet dataset and up to 4.7\% higher accuracy on various downstream tasks, while still maintaining the original TNNs' inference efficiency.
\end{itemize}

\section{Related Works}
\subsection{Tiny Neural Network} 

Tiny deep learning aims to develop TNNs with reduced network sizes, lower memory and computational costs, and acceptable accuracy, enabling deep learning-powered solutions in resource-constrained IoT devices.
Existing techniques towards fulfilling the goal of tiny deep learning can mostly be categorized into two trends.
One trend is to design novel TNN architectures by leveraging either human expertise~\cite{sandler2018mobilenetv2} or automated tools, e.g., neural architecture search~\cite{cai2018proxylessnas}.
The other trend is to make use of compression techniques, including pruning~\cite{liu2017learning}, quantization~\cite{fu2021double}, dynamic inference~\cite{yu2021mia}, to further reduce network complexity on top of existing TNN architectures.


In this work, we propose to pursue a tiny deep learning solution that boosts TNNs' accuracy-efficiency trade-off from an underexplored and orthogonal direction: how to train TNNs to unleash their achievable accuracy more effectively.
To the best of our knowledge, the only pioneering work that focuses on a similar direction is NetAug~\cite{cai2021network}, which proposes to augment TNNs from the width dimension by introducing a wider supernet to assist training and then directly remove the supernet during inference. In contrast, NetBooster proposes to first expand a TNN \textit{from both the depth and width dimensions} to create a competent deep giant during TNN training, and then gradually contract it back to the original structure, \textit{instead of directly removing expanded parts}, to avoid unrecoverable information loss that could result in nontrivial accuracy drops.

\subsection{Data Augmentation and Regularization}


Data augmentation and regularization techniques have been proposed to alleviate the over-fitting issue associated with large-scale DNNs in order to boost their network generalization ability and thus the achievable accuracy. Specifically, data augmentation techniques focus on manipulating the input data samples~\cite{cubuk2018autoaugment}, while regularization techniques focus on the network aspect and randomly drop different components from the network~\cite{srivastava2014dropout} during training. 

However, a recent study~\cite{cai2021network} and Fig.~\ref{fig:fig1} (a) have shown that TNN training suffers from under-fitting instead of over-fitting. As a result, existing data augmentation and regularization techniques are unable to fully unleash the potential of TNNs.

\subsection{Knowledge Distillation}


Knowledge distillation (KD) aims to transfer the already learned knowledge from a larger teacher network to a smaller student network~\cite{hinton2015distilling}.
Instead of relying on a teacher network to provide guidance during training, NetBooster aims to inherit the learned features from the expanded deep giants to achieve higher accuracy than the original TNNs. As such, our proposed NetBooster is orthogonal to KD, and when combined with KD, it is expected to further enhance the performance of TNNs.

\subsection{Transfer Learning}

Motivated by DNNs' strong feature extraction capability, transfer learning~\cite{donahue2014decaf} has become a widely used paradigm for transferring knowledge across different domains~\cite{mormont2018comparison}. Despite the extensive efforts to boost transferability, larger pretrained networks are commonly believed to have better transferability due to their representative and generalizable features. Our proposed NetBooster aims to empower TNNs with high-quality features and thus better transferability by leveraging TNNs' corresponding deep giants through compound network augmentation. 
\section{The Proposed NetBooster Framework}

\subsection{Motivations and Inspirations}
\label{sec:method-overview}

\textbf{The key challenge for TNN training.} Due to the lack of sufficient network capacity, TNNs tend to suffer from severe under-fitting issues when being trained on large-scale datasets (e.g., ImageNet), limiting their ability to learn complex but representative features~\cite{cai2021network} and further hindering their achievable accuracy on downstream tasks.

\textbf{Inspirations for our works.}
Recent works have demonstrated that overparameterization during training can lead to improved achievable accuracy, while network complexity during inference can be reduced without adversely affecting accuracy~\cite{liu2021we}. Various techniques for compressing DNNs have also supported this observation. For example, the pruning method~\cite{liu2017learning} preserves the original dense network during training and then removes redundant neurons for inference.
Nevertheless, the aforementioned methods only focus on overparameterization from the width dimension. At the same time, existing work~\cite{nguyen2020wide} shows that DNNs with varying depth and width tend to learn different features, urging the need to introduce overparameterization into both dimensions.
Thus, if we can equip a given TNN (e.g., original TNN) with comprehensive overparameterization during training, and then restore the original TNN's structure during inference, the under-fitting issue can be effectively mitigated to achieve more accurate yet efficient TNN inference.
This has inspired us to design a principled expansion-then-contraction methodology by first expanding the original TNN to a more overparameterized network (e.g., deep giant) for better feature learning and then contracting it back to the original structure to preserve its efficiency.

\begin{figure}[!t]
    \centering
    \includegraphics[width=\linewidth]{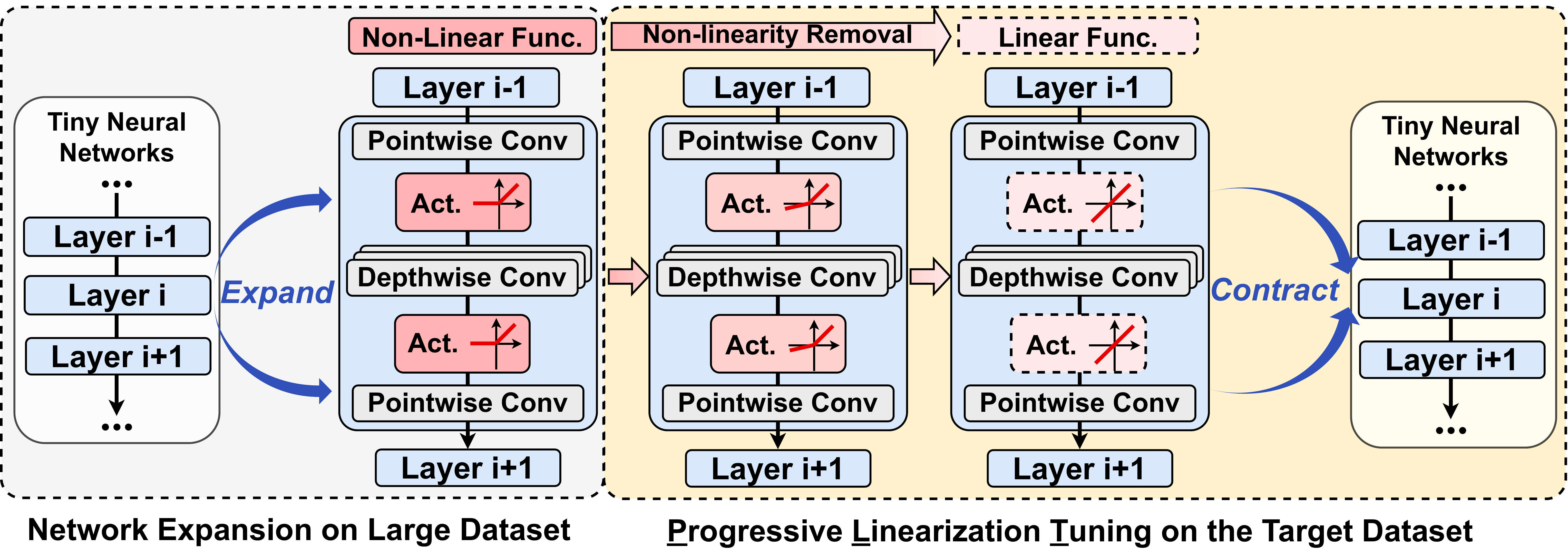}
    \vspace{-1.2em}
    \caption{An overview of the proposed NetBooster framework. In \textbf{NetBooster}, we augment the original TNN from both depth and width dimensions. Specifically, we uniformly select layers from the original TNN and expand them into inverted residual blocks~\cite{sandler2018mobilenetv2} to formulate the deep giant, helping to learn complex features. Then in PLT, we progressively decay the non-linear activation functions within the expanded inverted residual blocks to an identity mapping function and contract the expanded blocks back to the corresponding layers to maintain the original TNN's structure and inference efficiency. }
    \label{fig:framework}
    \vspace{-1.2em}
\end{figure}

\subsection{Overview}
\label{sec:method-overview}
\textbf{Implementation of expansion-then-contraction.}
Given the expansion-then-contraction principle, there are several potential implementations. Inspired by the success of RepVGG~\cite{ding2021repvgg}, which shows that parallel branches can be merged thanks to their linearity, we hypothesize that if we can properly remove the non-linear activation functions between layers, the consecutive layers can also be merged via a linear combination. 
Fortunately, recent works show that some of the activation functions can be safely removed from the network without hurting the task accuracy~\cite{jha2021deepreduce}. 
This motivates us to propose our expansion-then-contraction-based NetBooster training framework, which adopts two steps: \textbf{Step-1: Network Expansion}, where we augment the original TNN from both depth and width dimensions during training to construct the corresponding deep giant by replacing some layers in the original TNN with multi-layer blocks, aiming to increase the original TNN's capacity and alleviate its under-fitting issue during training and thus enabling a better feature learning on the large-scale training dataset, and \textbf{Step-2: Progressive Linearization Tuning (PLT)}, where we progressively remove the non-linearity \textit{inside} the expanded blocks on the target dataset. After the non-linear layers inside the blocks are removed, we contract the expanded deep giant back to the original TNN at the end of training to inherit the learned features while ensuring the boosted accuracy does not come with additional inference overhead.

\textbf{Technical challenges to achieve NetBooster.}
While the aforementioned principle sounds straightforward, implementing such a training pipeline is non-trivial. In \textbf{Step-1}, naively expanding all layers with a high expansion ratio can lead to an excessively large network which is difficult to train. To enable a practical and effective expansion strategy, at least the following three questions need to be addressed: \uline{$\mathcal{Q}1$: what blocks to insert?} What kind of blocks we should use to expand the original TNN, \uline{$\mathcal{Q}2$: where to expand?} How to find a set of target layers to be expanded, and \uline{$\mathcal{Q}3$: how to determine the expansion ratio?} To what extent we should expand the target layers. 
In \textbf{Step-2}, how to contract the deep giant expanded from both width and depth dimensions while preserving the learned knowledge of the deep giant is still an open question. Despite the method proposed in \cite{ding2021repvgg} can merge the parallel convolution layers into one single layer via linear combination, the non-linear activation layers between sequentially connected convolution layers make it impossible to merge convolution layers along the depth dimension directly.
We next elaborate on our proposed solutions to tackle the above challenges and the design of each step.

\subsection{Step 1: Network Expansion}
\label{sec:method-expansion}
The network expansion step aims to increase the capacity of the original TNN, transforming them into a more powerful deep giant. This boosts their ability to learn complex and representative features from large-scale training datasets, improving task accuracy and transferability.
To answer the questions raised in Sec.~\ref{sec:method-overview}, we propose the following criteria when expanding the network:
\begin{enumerate}[label=\textbf{\alph*.}]
    \item \textbf{Structural consistency:} To guarantee that applying NetBooster does not change the original TNN's structure for inference, each expansion block needs to be able to be contracted back to the original single layer via linear combination in the PLT step. Thus, for the network expansion step, \uline{the receptive field of each expansion block should be equal to that of the original layer}. 
    
    \item\textbf{Sufficient capacity:} Motivated by the findings in~\cite{cai2021network}, we aim to alleviate the under-fitting issue and ease the learning process by creating a deep giant with increased capacity. Thus, \uline{we should sufficiently expand the original TNN from multiple positions and dimensions} (i.e., expand width with increased expansion ratios and depth by inserting multiple layers).  

    \item \textbf{{Effective feature inheritance:}} In addition to having sufficient capacity, it is equally important to effectively inherit the learned features of the deep giant. As suggested in~\cite{mirzadeh2020improved}, excessively large networks tend to learn significantly different feature distribution from that of small networks, which can not only forbid small networks from inheriting but even hurt small networks' task accuracy. Thus, \uline{(1) the complexity gap between the original TNN and its expanded deep giant should not be too large and (2) the selected layers to be expanded should contain sufficient parameters} to ensure an effective knowledge inheritance from the deep giant. 
\end{enumerate}

Based on the above criteria, we answer the questions raised in Sec.~\ref{sec:method-overview} below:

\uline{$\mathcal{Q}1$. What kind of block to use?} We select the type of inserted blocks from a pool of well-established DNN building blocks (e.g., the basic and bottleneck blocks in ResNet~\cite{he2016deep} and the inverted residual block in MobileNetV2~\cite{sandler2018mobilenetv2}). To maintain structure consistency (criteria \textbf{a.}), we eliminate the basic block as it stacks two layers with large convolution kernels, leading to a receptive field larger than that of the original layer. To narrow down the complexity gap for effective feature inheritance (criteria \textbf{c.}), we select the inverted residual block over the bottleneck block.

\uline{$\mathcal{Q}2$. Where to expand?} The achievable accuracy of NetBooster is limited by a trade-off between increasing the network capacity by constructing a larger deep giant (criteria \textbf{b.}) and improving the feature inheritance effectiveness by narrowing down the capacity gap between the deep giant and original TNN (criteria \textbf{c.}). A simple but effective way to push the aforementioned trade-off further is to consider the knowledge inheritance effectiveness from a more fine-grained granularity (i.e., layer-wise instead of model-wise). Specifically, multiple layers can have a better representation ability than a single layer. Thus, the expanded block's learned complex features can be more effectively inherited by distributing them to multiple adjacent layers in the original TNN. To this end, we propose to uniformly select layers to be expanded from the original TNN, which can guarantee that there are sufficient layers to inherit learned features from each of the expanded blocks.

\uline{$\mathcal{Q}3$. How to determine the expansion ratio?} Similar to $\mathcal{Q}2$, the selection of the adopted expansion ratio has to trade-off between the network capacity (criteria \textbf{b.}) and the effectiveness of knowledge inheritance (criteria \textbf{c.}). However, thanks to the proposed uniform expansion strategy in $\mathcal{Q}2$, we empirically find that the commonly used expansion ratio, 6~\cite{sandler2018mobilenetv2} in the inserted inverted residual blocks works well on balancing the aforementioned two criteria.

\subsection{Step 2: Progressive Linearization Tuning (PLT)}
\label{sec:method-contraction}

The next step is to recover the original TNN's structure on the target dataset while inheriting the knowledge learned by the deep giant. Inspired by~\cite{ding2021repvgg}, which proposes to merge parallel convolution layers into one single layer, we find that sequentially connected layers can also be merged via linear combinations by properly removing the non-linear operations between them. To achieve this, we propose \textbf{PLT} to progressively remove the non-linearity from the expanded deep giant and then contract it back to the original TNN during finetuning on the target dataset. 

\noindent \textbf{Motivating observation. }
Non-linearity has been considered a key enabler for the promising performance of DNNs and most existing works use the combination of convolution and non-linear activation layers as a basic design unit. In parallel, recent works~\cite{jha2021deepreduce} have shown that \textbf{non-linearity within DNNs can be highly redundant for inference}, a large portion of element-wise non-linear activation functions can be removed from a DNN, and the complex features learned from the original TNN during training can be largely preserved. Inspired by the revolution from element-wise pruning to structure pruning, \textbf{we aim to step further and remove the non-linearity in a structured manner (i.e., layer-wisely)}.

\noindent \textbf{Non-linearity removal.}
We propose to transform the expanded deep giant back to the original TNN meanwhile preserve the learned features by slowly decaying the non-linear activation functions.

Without loss of generality, we take the ReLU activation function as an example as it is the most commonly adopted activation function in TNNs, and the following discussion can also be extended to other activation functions like ReLU6. Here the ReLU function is defined as:
\begin{equation}
    Y_l = \max(0, X_l),
\end{equation}
where $X_l$ and $Y_l$ are the input and output of layer $l$, respectively. We change the formulation of ReLU to the following format, 
\begin{equation}
    Y_l = \max(\alpha_l X_l, X_l),
    \label{eq:reformulate_relu}
\end{equation}
where $0<\alpha_l<1$ is the slope parameter to manipulate the non-linearity of the corresponding activation layer. When $\alpha_l=0$, it is exactly the ReLU function. When $\alpha_l=1$, the activation function is decayed to an identity mapping. 

Given a list $L$ of non-linear activation layers to be removed, we gradually increase $\alpha_{l'}$ for $l'\in L$ from 0 to 1 in $E_d$ epochs, the value of $\alpha_{l'}$ is uniformly increased in each iteration. When $\alpha_{l'}=1$, Eq.~\ref{eq:reformulate_relu} is an identity mapping, and thus the non-linearity is removed. 

\noindent \textbf{Expanded block contraction.}
With the non-linear activation layers removed, the remaining layers can be contracted into one layer via simple linear combinations. 

\noindent \uline{Formulation:}
Without loss of generality, we take two convolution layers as an example. Given the input to the first layer $X\in\mathcal{R}^{h_1\times w_1\times c_1}$, the output $Y\in\mathcal{R}^{h_3\times w_3\times c_3}$, as well as the kernels of 
two layers $K^1\in\mathcal{R}^{k_1\times k_1\times c_1\times c_2}$ and $K^2\in\mathcal{R}^{k_2\times k_2\times c_2\times c_3}$, where $k_{1/2}$ are the kernel sizes, $h_{1/2/3}, w_{1/2/3}$ are the heights and widths of corresponding feature maps, $c_{1/2/3}$ are the channel numbers. The overall functionality of the two convolution layers can be formulated as
\begin{align} 
    \begin{split} \label{eq:conv1}
 & Y_{p,q,o} = \sum^{k-1}_{i=0}\sum^{k-1}_{j=0}\sum^{c_1-1}_{m=0}X_{p-i, q-j,m}K_{i,j,m,o}, \\ 
    \end{split} \\
    \begin{split} \label{eq:conv2}
where & \,\, K_{i,j,m,o} = \sum_{s=s_l}^{s_h}\sum_{t=t_l}^{t_h}\sum_{n=0}^{c_2-1}K^1_{i-s, j-t, m,n}K^2_{s,t,n,o},
    \end{split} 
\end{align}

\noindent where $k=k_1+k_2-1$, $s_l=\max(0, i-k_1+1)$, $s_h=\min(k_2-1, i)$, $t_l=\max(0, j-k_2+1)$ and $t_h=\min(k_2-1, j)$. 

\noindent\uline{Remark:}
It is worth noting that different expansion ratios of the inserted inverted residual block will result in the same computational cost after contraction since the input and output channels after contraction are always equal to the input channel of the first layer and the output channel of the last layer, respectively, regardless of intermediate channel numbers (i.e., $c_2$).
\begin{table}[t]
    \centering
    \vspace{-1em}
    \caption{Benchmarking on \textbf{ImageNet}. 'r' is the input resolution.}
    \vspace{-0.8em}
    \resizebox{0.9\linewidth}{!}{
    \begin{tabular}{c|cc|cc}
    \toprule
        Network  & FLOPs & Params & Training Method & Accuracy \\
        \midrule
        \multirow{7}{*}{\makecell[c]{MobileNetV2-Tiny\\(r=144)}} & \multirow{7}{*}{23.5M} & \multirow{7}{*}{0.75M} & Vanilla & 51.2 \\
        &  &  & RocketLaunch~\cite{zhou2018rocket} & 51.8 \\
        &  &  & tf-KD~\cite{yuan2020revisiting} & 51.9 \\
        &  &  & RCO-KD~\cite{jin2019knowledge} & 52.6 \\
        &  &  & NetAug~\cite{cai2021network} & 53.0 \\
        &  &  & NetBooster &\textbf{53.7}\\
        \midrule
        \multirow{3}{*}{\makecell[c]{MCUNet\\(r=176)}} & \multirow{3}{*}{81.8M} & \multirow{3}{*}{0.74M} & Vanilla & 61.4 \\
        &  &  & NetAug~\cite{cai2021network} & 62.5 \\
        &  &  & NetBooster &\textbf{62.8}\\
        \midrule
        \multirow{3}{*}{\makecell[c]{MobileNetV2-50\\(r=160)}} & \multirow{3}{*}{50.2M} & \multirow{3}{*}{1.95M} & Vanilla & 61.4\\
        &  &  & NetAug~\cite{cai2021network} & 62.5\\
        &  &  & NetBooster & \textbf{62.7}\\
        \midrule
        \multirow{3}{*}{\makecell[c]{MobileNetV2-100\\(r=160)}} & \multirow{3}{*}{154.1M} & \multirow{3}{*}{3.47M} & Vanilla & 69.6\\
        &  &  & NetAug~\cite{cai2021network} & 70.5 \\
        &  &  & NetBooster & \textbf{70.9} \\
        \bottomrule
    \end{tabular}
    }
    \vspace{-2.2em}
    \label{tab:imagenet}
\end{table}

\section{Experiments}
\subsection{Experiments Setup}
\textbf{Tasks, datasets, and networks} We consider \uline{two tasks}, including image classification and object detection, with \uline{seven datasets} to provide a thorough evaluation of NetBooster. Specifically, to evaluate NetBooster's performance in alleviating the under-fitting issue to achieve a higher accuracy on the large-scale dataset, we consider the ImageNet dataset. To evaluate how deep giant's learned representation helps with downstream tasks, we consider image classification tasks on five datasets, including CIFAR-100, Cars~\cite{krause20133d}, Flowers102~\cite{nilsback2008automated}, Food101~\cite{bossard2014food}, and Pets~\cite{parkhi2012cats}. We also evaluate NetBooster on the downstream object detection task with the Pascal VOC dataset~\cite{everingham2010pascal}. We consider \uline{four networks}, including MobileNetV2-100/50/Tiny~\cite{sandler2018mobilenetv2}, and a neural architecture searched hardware friendly network MCUNet~\cite{lin2020mcunet}.

\textbf{Baselines.}  We benchmark the proposed NetBooster over five baselines including networks trained with standard vanilla training, a series of SOTA KD algorithms (i.e., tf-KD~\cite{yuan2020revisiting}, RCO-KD~\cite{jin2019knowledge}, and RocketLaunch~\cite{zhou2018rocket}), and NetAug~\cite{cai2021network}, which is a pioneering work in boosting TNN training performance. 

\textbf{Expansion strategy.} We uniformly expand 50\% of blocks in the original TNN. To expand each block, we replace the first pointwise convolution layer with an inverted residual block with an expansion ratio of 6. The kernel size of the depthwise convolution layer in the inserted inverted residual block is set to 1 to make the inserted block's receptive field the same as the pointwise convolution. 

\textbf{Training settings.} We develop our training settings based on the commonly adopted settings. Specifically, when evaluating \uline{NetBooster's performance in improving TNNs' accuracy on ImageNet dataset}, we follow~\cite{cai2021network} to train the deep giant for 160 epochs using SGD optimizer with a batch size of 1024, an initial learning rate of 0.2 and cosine anneal learning rate schedule. In PLT, we set $E_d=40$, and further finetune for 110 epochs. For \uline{downstream tasks}, we use the ImageNet pretrained deep giant as the starting point and develop our training recipe on CIFAR-100 based on~\cite{tian2019contrastive}, on Cars, Flowers102, Food101, and Pets based on~\cite{salman2020adversarially}, and on Pascal VOC based on~\cite{cai2021network}. In all experiments on downstream tasks, we assign $E_d$ to be 20\% of the total tuning epochs in PLT.

\subsection{Easing \uline{\textbf{\textit{Constraint 1}}}: Benchmarking on Large-scale Dataset}
To evaluate whether the proposed NetBooster can help TNNs to learn the complex features of the large-scale dataset and thus improve accuracy, we benchmark NetBooster on the ImageNet dataset with vanilla training, NetAug, and various KD algorithms.
As shown in Table~\ref{tab:imagenet}, NetBooster achieves 1.3\% $\sim$ 2.5\% accuracy improvements over the vanilla training, showing its strong ability in boosting the TNNs' accuracy by standing on the shoulder of deep giants generated by NetBooster.

Compared with the KD baselines, our proposed NetBooster achieves 0.9\% $\sim$ 1.1\% accuracy improvement without guidance from the teacher network (Assemble-ResNet50~\cite{lee2020compounding}), suggesting that Network Expansion in NetBooster can equip the deep giant with sufficient capacity to learn complex features at least comparable with the large teacher DNN used in the KD baselines and the learned features can be effectively inherited from the PLT step.  

Compared with NetAug, which is a pioneering work focusing on a similar scenario as NetBooster, NetBooster also achieves superior accuracy over NetAug, suggesting the multi-dimensional Network Expansion and the PTL for features inheritance is more effective than the network width expansion and directly dropping augmented neuron adopted in NetAug. 

\begin{table}[t]
    \centering
    \vspace{-1em}
    \caption{Benchmarking on \textbf{downstream image classification datasets}. 'r' is the input resolution.}
    \vspace{-0.5em}
    \resizebox{0.99\linewidth}{!}{
    \begin{tabular}{c|c|ccccc}
        \toprule
        Network & Training Method & CIFAR-100 & Cars & Flowers102 & Food101 & Pets \\ 
        \midrule
         \multirow{2}{*}{\makecell[c]{MobileNetV2-Tiny\\(r=144)}} & Vanilla & 74.07 & 76.18 & 90.01 & 75.43 & 78.30 \\
         & NetBooster  & 75.46 & 80.93 & 90.53 & 75.96 & 78.90 \\
         \midrule
        \multirow{4}{*}{\makecell[c]{MobileNetV2-35\\(r=160)}} & Vanilla & 76.08 & 78.36 & 90.63 & 76.80 & 80.64\\
         & Vanilla + KD & 76.38 & 77.47 & 91.41 & 77.02 & 82.44 \\
         & NetBooster  & 76.66 & 80.91 & 91.16 & 77.26 & 80.92\\
         & NetBooster + KD & 77.15 & 83.36 & 92.68 & 77.81 & 83.37 \\
        \bottomrule
    \end{tabular}
    }
    \vspace{-0.8em}
    \label{tab:finegrain}
\end{table}

\begin{table}[t]
    \centering
    \caption{Benchmarking on \textbf{object detection} tasks with Pascal VOC dataset with MobileNetV2-35 at 416 resolution.}
    \vspace{-0.5em}
    \resizebox{0.6\linewidth}{!}{
    \begin{tabular}{c|ccc}
    \toprule
        Method & Vanilla & NetAug & NetBooster \\
        \midrule
        AP50 & 60.8 & 62.4 & \textbf{62.6}\\
        \bottomrule
    \end{tabular}
    }
    \label{tab:det}
    \vspace{-2.2em}
\end{table}

\subsection{Easing \uline{\textbf{\textit{Constraint 2}}}: Benchmarking on Downstream Tasks}
To evaluate whether the learned complex and representative features in the deep giant from the large-scale dataset can further help the original TNN to achieve better accuracy on downstream tasks, we first evaluate NetBooster's performance when transferring the ImageNet pretrained deep giant to five representative downstream image classification datasets with PLT. As shown in Table~\ref{tab:finegrain}, compared with vanilla training, NetBooster achieves 0.46\% $\sim$ 4.75\% accuracy improvement, showing that the features learned by the deep giant are effectively inherited after PLT. It is worth noting that NetBooster is also orthogonal to KD, applying KD on top of NetBooster can lead to another 0.49\% $\sim$ 2.45\% accuracy boost over using NetBooster alone. 

We further evaluate NetBooster's performance when transferring to the Pascal VOC object detection task. As shown in Table~\ref{tab:det}, NetBooster achieves 1.8 and 0.2 higher AP50 compared with vanilla training and NetAug, respectively. This proves that NetBooster can be considered as a general method to boost TNNs' performance across various tasks.

\begin{table}[t]
    \centering
    \vspace{-1em}
    \caption{Ablation study on \textbf{what kind of block} to insert.}
    \vspace{-0.5em}
    \resizebox{0.8\linewidth}{!}{
    \begin{tabular}{c|ccc}
    \toprule
    Inserted Block Type & Expanded Acc. & Final Acc. \\
    \midrule 
    Vanilla & - & 51.20 \\ 
    \midrule
        Inverted Residual & 54.90 & 53.70\\ 
        Basic Block & 54.52 & 53.41 \\
        Bottleneck Block & 55.23 & 53.62 \\
        \bottomrule
    \end{tabular}}
    \label{tab:block to insert}
    \vspace{-1.2em}
\end{table}

\subsection{Validating Expansion Strategy}
We validate our answer to each question in Sec.~\ref{sec:method-expansion} by validating the impact of replacing our proposed strategy with alternatives when training MobileNet-Tiny on the ImageNet dataset with an input resolution of 144.

\uline{$\mathcal{Q}1$. What kind of block to use}: We ablate the impact of expanding with different kinds of blocks and report the results in Table~\ref{tab:block to insert}. Expanding with inverted residual blocks leads to slightly better results (0.08\% $\sim$ 0.29\%). It shows (1) the NetBooster framework can robustly boost TNNs' performance, and (2) inserting inverted residual blocks is an effective choice. 

\uline{$\mathcal{Q}2$. Where to expand}: We ablate different expansion locations' impacts and report our findings in Table~\ref{tab:block to expand}. We observe that uniformly expanding the model achieves a 1.08\% $\sim$ 2.20\% higher accuracy compared with excessively expanding the first/middle/last part of the network, proving the necessity to expand the model uniformly.

\uline{$\mathcal{Q}3$. How to determine expansion ratio}: We ablate different selections of expansion ratios in the inserted inverted residual blocks and report the results in Table.~\ref{tab:expansion ratio}. We observe that NetBooster with commonly used expansion ratios (i.e., 4 $\sim$ 6) consistently improves the TNNs' accuracy, further proving NetBooster's robustness to hyperparameter selection. 

\begin{table}[t]
    \centering
    \caption{Ablation study on \textbf{which block} to expand. }
    \vspace{-0.5em}
    \resizebox{0.99\linewidth}{!}{
    \begin{tabular}{c|cc|cc}
    \toprule
        \multirow{2}{*}{Expansion} & \multicolumn{2}{c|}{Expanded} &\multirow{2}{*}{Expanded Acc.} & \multirow{2}{*}{Final Acc.}  \\
        & FLOPs & Params & & \\
        \midrule
        Vanilla & 29.4M & 0.75M & - & 51.20 \\
        \midrule
        Expand First 8 & 65.0M & 0.83M & 51.46 & 51.50 \\
        Expand Middle 8 & 49.6M & 0.93M & 52.98 & 52.62 \\
        Expand Last 8 & 51.2M & 1.25M & 53.90 & 52.47 \\
        \midrule
        Uniform Expand 8 & 63.9M & 0.99M & 54.90 & 53.70 \\
        \bottomrule
    \end{tabular}
    }
    \vspace{-1.2em}
    \label{tab:block to expand}
\end{table}
\begin{table}[t]
    \vspace{-1em}
    \centering
        \caption{Ablation study on the \textbf{expansion ratio}.}
    \vspace{-0.6em}
    \resizebox{0.7\linewidth}{!}{
        \begin{tabular}{c|cccc}
        \toprule
            Expansion ratio & 2 & 4 & 6 & 8 \\
            \midrule
            Final Acc. & 52.94 & 53.52 & 53.70 & 52.56\\
            \bottomrule
        \end{tabular}
        }
    \label{tab:expansion ratio}
    \vspace{-2.6em}
\end{table}

\section{Conclusion}

In this paper, we discover and promote a new paradigm for training TNNs to empower their achievable accuracy via augmenting both dimensions of TNNs (i.e., depth and width) during training. Furthermore, we propose a framework dubbed NetBooster, which is dedicated to boosting the accuracy of SOTA TNNs by using an expand-then-contract training strategy to alleviate TNNs' under-fitting issues. Finally, we make heuristic efforts to empirically explore what/when/where to augment when training TNNs with our proposed NetBooster. 
Extensive experiments show that NetBooster consistently leads to a nontrivial accuracy boost (e.g., 1.3\% $\sim$ 2.5\%) on top of SOTA TNNs on ImageNet and as much as 4.7\% higher accuracy on various downstream tasks, while maintaining their inference efficiency.


\bibliographystyle{IEEEtranS}
\bibliography{ref}


\end{document}